\documentclass[11pt, conference, letterpaper]{IEEEtran}

\usepackage{amsmath,amssymb,amsfonts}

\usepackage{mathtools}  
\usepackage{tabulary}
\usepackage{booktabs}

\usepackage[top=0.60in, left=0.65in, bottom=.60in, right=0.7in]{geometry}

\usepackage[font=small]{caption}
\usepackage{subcaption}
\usepackage[numbers, compress]{natbib}

\usepackage[hyphens]{url}
\usepackage[hidelinks]{hyperref}

\usepackage[ruled,vlined,noresetcount]{algorithm2e}

\usepackage[utf8]{inputenc}
\usepackage{authblk}
\SetKw{KwBy}{by}

\title{A new Potential-Based Reward Shaping for Reinforcement Learning Agent}

\author[1]{Babak Badnava \thanks{babak.badnava@ku.edu}}
\author[2]{Mona Esmaeili \thanks{mesmaeili@unm.edu}}
\author[3]{Nasser Mozayani\thanks{mozayani@iust.ac.ir}}
\author[2]{Payman Zarkesh-Ha  \thanks{pzarkesh@unm.edu}}
\affil[1]{University of Kansas, Lawrence, KS 66045, USA.}
\affil[2]{University of New Mexico, Albuquerque, NM 87131, USA.}
\affil[3]{Iran University of Science and Technology, Tehran, 16846, Iran.}

\makeatletter
\def\ps@headings{
\let\@oddhead\@empty
\let\@evenhead\@empty
\def\@oddfoot{\@IEEEheaderstyle\hfil\thepage}%
\def\@evenfoot{\@IEEEheaderstyle\thepage\hfil\hbox{}}
}
\def\ps@IEEEtitlepagestyle{
\def\@oddhead{\footnotesize This work is presented as a conference paper in 2023 IEEE 13th Annual Computing and Communication Workshop and Conference (CCWC). \hfill}%
\let\@evenhead\@empty
\def\@oddfoot{\footnotesize 
This work was done during the time that Babak Badnava was attending the Iran University of Science and Technology \hfill}%
\let\@evenfoot\@empty
}
\makeatother

\begin{document}

\maketitle

\IEEEpubidadjcol
\begin{abstract}
Potential-based reward shaping (PBRS) is a particular category of machine learning methods that aims to improve the learning speed of a reinforcement learning agent by extracting and utilizing extra knowledge while performing a task. There are two steps in the transfer learning process: extracting knowledge from previously learned tasks and transferring that knowledge to use it in a target task. The latter step is well discussed in the literature, with various methods being proposed for it, while the former has been explored less. With this in mind, the type of knowledge that is transmitted is very important and can lead to considerable improvement.
Among the literature of both transfer learning and potential-based reward shaping, a subject that has never been addressed is the knowledge gathered during the learning process itself.
In this paper, we presented a novel potential-based reward shaping method that attempted to extract knowledge from the learning process. The proposed method extracts knowledge from episodes' cumulative rewards. The proposed method has been evaluated in the Arcade learning environment, and the results indicate an improvement in the learning process in both the single-task and the multi-task reinforcement learner agents.

\end{abstract}


\section{Introduction}
In reinforcement learning (RL) problems, an agent learns to maximize a reward signal, which may be time-delayed \cite{Sutton98}. The RL framework has gained much success in handling complex problems in last years. However, mastering difficult tasks is often slow. Thus most of the RL researcher focuses on improving the speed of the learning process by using the knowledge that provided by an expert or a heuristic function.

Transfer learning (TL) is one of the approaches that try to improve the speed of learning by using knowledge extracted from previously learned tasks \cite{initial}. Reward shaping (RS) is also one of the methods that have been used for transferring such knowledge or any other type of knowledge.
 
In problems that could be modeled using RL, many tasks have kind of sparse reward signal. For example, there are some tasks that an agent will receive nothing until it gets to a goal or it gets nothing until some events occur in the environment. The reward function of a task in an environment is independent of time, and the agent tries to maximize the sum of its discounted reward in an episode, while from an episode to another episode of learning, the agent could use a knowledge extracted from past episodes to reinforce the reward signal. There is much information that can be used to reinforce the reward signal. For example, after one episode of learning has done, the agent could compare its performance with the best episode of the learning or the worst one, and by this comparison, it could improve the learning process of itself.
\section{Background}
\subsection{Reinforcement learning}
RL is a group of machine learning methods, which can be used for training an agent using environmental feedback. An RL agent in an environment can be modeled as a Markov decision process(MDP) \cite{Sutton98}. An MDP is a tuple like $(S, A, T, \gamma, R)$, and in this tuple: $S$ is a set of states that appear in the environment; $A$ is a set of agent's possible actions; $T : S \times A \times S \rightarrow [0,1]$  is a function which gives probability of reaching state $s^{'}$ while agent currently is in state $s$ and chose action $a$ to perform; $R : S \times A \times S \rightarrow \mathbb{R}$ is a function which gives amount of reward the agent will receives when performed the action $a$ in state $s$ and transferred to state $s^{'}$; $\gamma$ called discount factor which indicates how future rewards are important for the agent.
\begin{equation}\label{eq:return} \small
G_{t} = \sum_{k=0}^{\infty} \gamma^{k} R_{t+k+1}
\end{equation}

The agent wants to maximize equation \ref{eq:return}, which is called the expected discounted return, during its lifetime \cite{Sutton98}. There are many methods for training RL agents. Methods such as Q-learning\cite{q-learning92} and SARSA which estimate a value for each $(state, action)$ pair and by using this estimated value compute a policy that can maximize the agent's discounted return. There are also some methods that use policy gradient \cite{reinforce92,sutton-policy-grad,policy-grad} to approximate a policy and use this policy for the agent's decision making. 

\begin{equation}\label{eq:q-lear}
\resizebox{0.91\hsize}{!}{
$Q(s_{t},a_{t}) \leftarrow Q(s_{t},a_{t}) + \alpha \left[ R_{t+1} + \gamma \max_{a} Q(s_{t+1}, a) - Q(s_{t},a_{t}) \right]$}
\end{equation}
\subsection{Reward shaping}
Reward shaping (RS) refers to allowing an agent to train on an artificial reward signal rather than environmental feedback \cite{taylor-survey}. However, this artificial signal must be a potential function. Otherwise, the optimal policy will be different for the new MDP \cite{policy-invar-ng}. Authors of \cite{policy-invar-ng} prove that $F$ is a potential function if there exists a real-valued function $\phi : S \rightarrow \mathbb{R}$ such that for all $s \in S - \lbrace s_{0} \rbrace, a \in A, s^{'} \in S$:
\begin{equation}\label{eq:p-static}
F(s,s^{'}) = \gamma \phi(s^{'}) - \phi(s)
\end{equation}
Consequently by using this potential function, the optimal policy in the new MDP($\mathcal{M^{'}}=(S,A,T,\gamma , R+F)$) remains optimal for the old MDP($\mathcal{M}=(S,A,T,\gamma, R)$). In the new MDP, equation \ref{eq:p-static}, and all of the extended equations from equation \ref{eq:p-static}, $\gamma$ is unchanged and must have the same value as in the old MDP. There are also other works in the literature that extend the potential function. Authors of \cite{princ-rs} have shown that the potential function can be a function of the joint state-action space. Next, Authors of \cite{dynamic-pbrs} have shown that the potential function could be a dynamic function and that might change over the time while all the properties of potential based reward shaping (PBRS) remain constant. In addition to \cite{princ-rs} and \cite{dynamic-pbrs}, \cite{arbitrary-rs} combines these two extensions and show that any function in the form of equation \ref{eq:p-sa-dynamic} could be a potential function.
\begin{equation}\label{eq:p-sa-static}
F(s,a,s^{'},a^{'}) = \gamma \phi(s^{'},a^{'}) - \phi(s,a)
\end{equation}
\begin{equation}\label{eq:p-dynamic}
F(s,t,s^{'},t^{'}) = \gamma \phi(s^{'},t^{'}) - \phi(s,t)
\end{equation}
\begin{equation}\label{eq:p-sa-dynamic}
F(s,a,t,s^{'},a^{'},t^{'}) = \gamma \phi(s^{'},a^{'},t^{'}) - \phi(s,a,t)
\end{equation}

By using reward shaping Q-learning update rule will turn into equation \ref{eq:q-l-rs}. In equation \ref{eq:q-l-rs}, $F$  could be any of equations represented in  \ref{eq:p-static}, \ref{eq:p-sa-static}, \ref{eq:p-dynamic}, or \ref{eq:p-sa-dynamic}. Equation \ref{eq:q-l-rs} will give us this opportunity to enhance the learning by providing some extra information about the problem for the agent. This additional knowledge can come from any source like human knowledge of the problem, some heuristic function, or the knowledge that extracted by the agent.

\begin{equation}\label{eq:q-l-rs}\resizebox{0.91\hsize}{!}{
$Q(s_{t},a_{t}) \leftarrow Q(s_{t},a_{t}) + \alpha \left[ R_{t+1} + F + \gamma \max_{a} Q(s_{t+1}, a) - Q(s_{t},a_{t}) \right]$
}
\end{equation}
    

\section{Related works}
The first method, which should be considered, is the Multi-grid RL with RS, \cite{multigrid-rs} which propose a method for learning a potential function in an online manner. Authors of \cite{multigrid-rs} estimate a value function during the learning process and by using this value function shape a potential function. The estimated value function is the value of an abstract state. State abstraction could be applied by any method.
\begin{equation}
V(z) = \left(1 - \alpha_{v}\right) V(z) + \alpha_{v} \left( r_{v} + \gamma_{v}^{t}(z^{'})\right)
\end{equation}
\begin{equation}
F(s,s^{'}) = \gamma_{v}V(z^{'}) - V(z)
\end{equation}
Another work that has been done in the literature is \cite{plan-based-rs}, which proposed a potential function based on a plan. In \cite{plan-based-rs}, the agent gets an extra reward based on progress in the plan. In equation \ref{eq:plan-rs}, z is an abstract state that can be any state of the plan, and the function $step(z)$ returns the time step at which given the abstract state $z$ appears during the process of executing the plan.
\begin{equation}\label{eq:plan-rs}
\phi(s) = step(z)
\end{equation}
Another work that is an extended version of  \cite{plan-based-rs}, for multi-agent reinforcement learning, is \cite{plan-rs-mas}. In \cite{plan-rs-mas} two methods have been proposed, which can be used to shape the reward signal of agents and improve their learning process.

Another work that has been presented in the literature of transfer learning is a method that transfers the agent's policy using reward shaping. Authors of \cite{policy-rs} assume that there is a mapping function, which maps target task to source task. By using this mapping function, they define a potential function for shaping the reward signal. The proposed potential function as shown in equation \ref{eq:policy-rs} is defined based on the policy of source task. In equation \ref{eq:policy-rs}, $\mathcal{X}_{S}$ is a mapping function from target task state space to source task state space, and $\mathcal{X}_{A}$ is a mapping function from target task action space to source task action space, and $\pi$ is the policy of agent in the source task.
\begin{equation}\label{eq:policy-rs}
\phi (s,a) = \pi (\mathcal{X}_{S}(s),\mathcal{X}_{A}(a))
\end{equation}
In \cite{policy-rs}, RS has been used as a knowledge-transfer procedure, and a learned policy from another task has been used as knowledge.
There are many other works in the literature that proposed a potential based reward shaping method for multi-agent reinforcement learning (MARL) and single-agent reinforcement learning (SARL). \cite{RLfD} proposed a method for RS, while there is a demonstration of a task. \cite{RLfD} proposed a method that uses a similarity measure for calculating the similarity between state-action pairs. Another work, which assumes a situation like that assumed in \cite{RLfD}, is \cite{RLfD-IRL}. Authors of \cite{RLfD-IRL} instead of using a similarity measure used an Inverse reinforcement learning approach to approximate a reward function for shaping the potential function. Another technique that is presented for MARL is \cite{diff-rs}. In \cite{diff-rs} difference reward used to shape a potential function. The difference reward helps the agent to learn the impact of its action on the environment by not considering the impact of other agents' action. 
\section{Proposed potential based reward shaping method}
Given these points and motivations, we go on to describe our approach to reinforce reward signal by using a knowledge gained from the learning process. So, we want a reward function that will change every time the agent make progress in the task. This reward function must encourage or punish the agent according to the best and worst episode until now. The reward function must also handle sparse reward signals. The function that represented in equation \ref{eq:p-timeVar} has properties that we want. As we know from \cite{policy-invar-ng}, changing reward function might change the optimal policy for the current task. Hence, we use reward shaping to manipulate the reward function in order to consider the agent's improvement during the learning process.
\begin{equation}\label{eq:p-timeVar}
\phi(s,a,t) = \Bigg\{ 
\begin{array}{ll}
	0 & R(s,a) = 0 \\
	1 + \frac{R^{ep} - R_{u}^{ep}(t)}{R_{u}^{ep}(t) - R_{l}^{ep}(t)} & O.W
\end{array}
\end{equation}
In equation \ref{eq:p-timeVar}: $R(s,a)$ is the immediate reward, which is a sparss reward signal; $R^{ep}$ is the sum of rewards in the current episode, which we call it $episode\,reward$; $R_{u}^{ep}(t)$ is the maximum value of $episode\,reward$ until now; and $R_{l}^{ep}(t)$ is the minimum value of $episode\,reward$ until now.
This function returns zero, while the reward signal of the environment has no information in it; the case to control the effect of sparse-reward. This function reinforces the reward signal by measuring a distance from a fixed point. We could use this approach with any kind of learning algorithm to boost the learning process. As presented in algorithm \ref{alg:pbrs-ai}, after each episode of learning the agent will change parameters of potential function if needed.

\begin{algorithm}
\caption{Q-learning + proposed PBRS method}\label{alg:pbrs-ai}
\SetAlgoLined
$\forall$ s,a Q(s,a) = 0 \\
s $\leftarrow$ statrt\,state \\
$max\,episode\,reward \leftarrow -\infty$ \\
$min\,episode\,reward \leftarrow \infty$ \\
$replay\, buffer \leftarrow null$ \\
 \While{True}{
 	$episode\,reward = 0$ \\
 	\While{s is not terminal}{
	 $a \leftarrow$	Choose\,an\,action\,based\,on\,a\,policy\,and\,given\,state \\
	 $r, s^{'} \leftarrow$ do\,action,\,get\,reward\,and\,observe\,next\,state \\
	 $episode\,reward \leftarrow episode\,reward + r$\\
	 \eIf{r = 0 }{
	 $\phi_{s} \leftarrow 0$ \\
	 }{
		$\phi_{s} \leftarrow 1 + \frac{episode\,reward - max\,episode\,reward}{max\,episode\,reward - min\,episode\,reward}$	 
	 }
	 append $replay\, buffer$ $(s,a,r,\phi_{s},s^{'})$ \\
	 s $\leftarrow s^{'}$\\
	 \If{update priode or s is terminal}{
	  \For{$i\gets$ length of replay buffer - 1 \KwTo $0$ \KwBy $-1$}{
		$s,a,r,\phi_{s},s^{'} \leftarrow$ $replay\,buffer$[i]	\\
				$\_,\_,\_,\phi_{s^{'}},\_ \leftarrow$ $replay\,buffer$[i+1]	\\
	  	 $F \leftarrow \gamma \phi_{s^{'}} - \phi_{s}$ \\
$	 Q(s,a) \leftarrow Q(s,a) + \alpha \left[ r + F + \gamma \max_{a} Q(s^{'}, a) - Q(s,a) \right]$ \\
  }

	$replay\, buffer \leftarrow null$ \\
	}

 	}
 	\uIf{$episode\,reward > max\,episode\,reward$}{
 	$max\,episode\,reward \leftarrow episode\,reward$\\
 	}
 	\ElseIf{$episode\,reward < min\,episode\,reward$}{
 	 	$min\,episode\,reward \leftarrow episode\,reward$\\
 	}	
}
\end{algorithm}


\begin{figure}[h]
	\centering
	\includegraphics[width=.5\textwidth, trim={.7cm .7cm .7cm .7cm},clip]{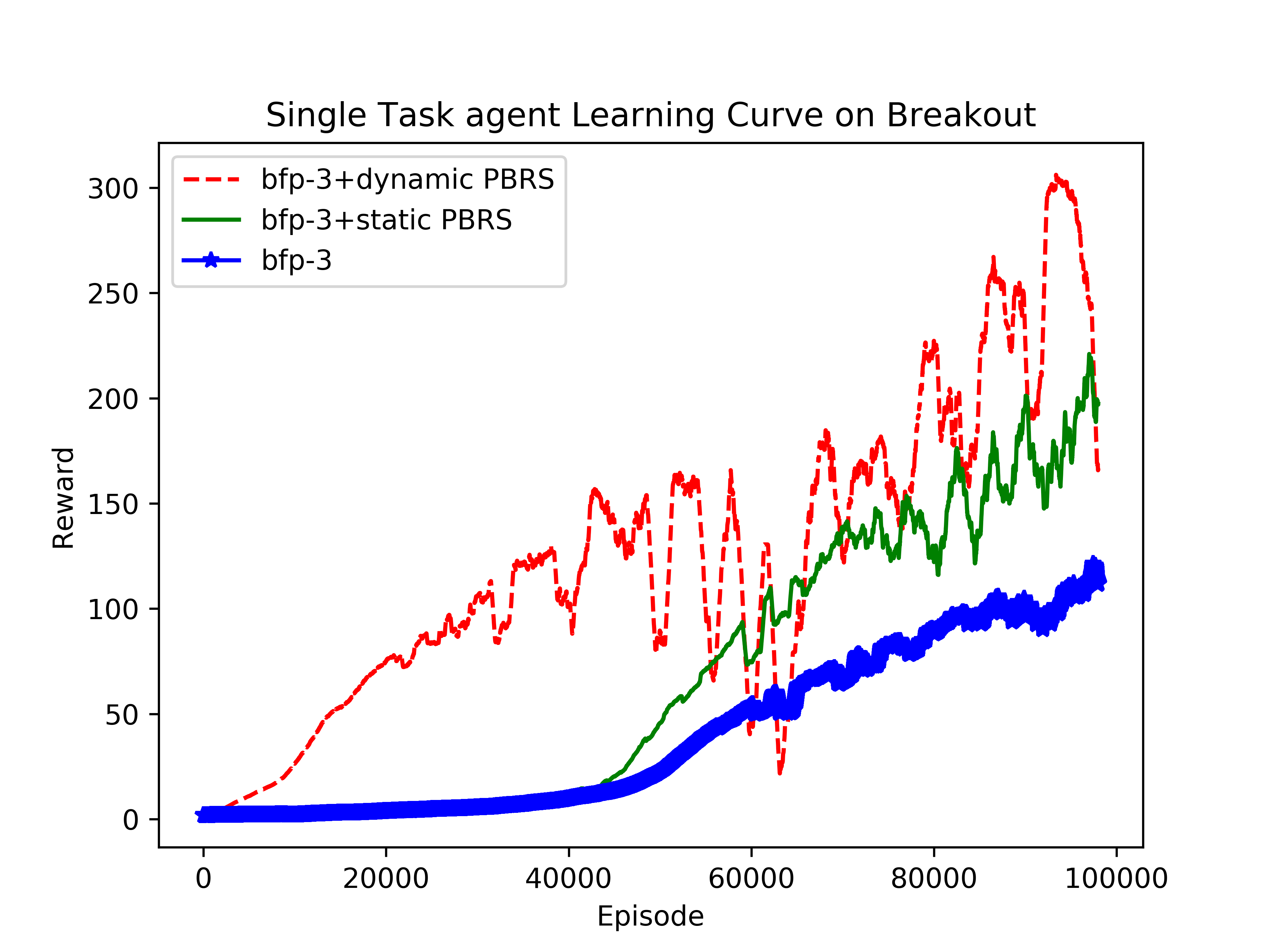}
\caption{single task agent learning curve on Breakout}
\label{fig:breakout-a}
\end{figure}

\begin{figure}[h]
	\centering
	\includegraphics[width=.5\textwidth, trim={.7cm .7cm .7cm .7cm},clip]{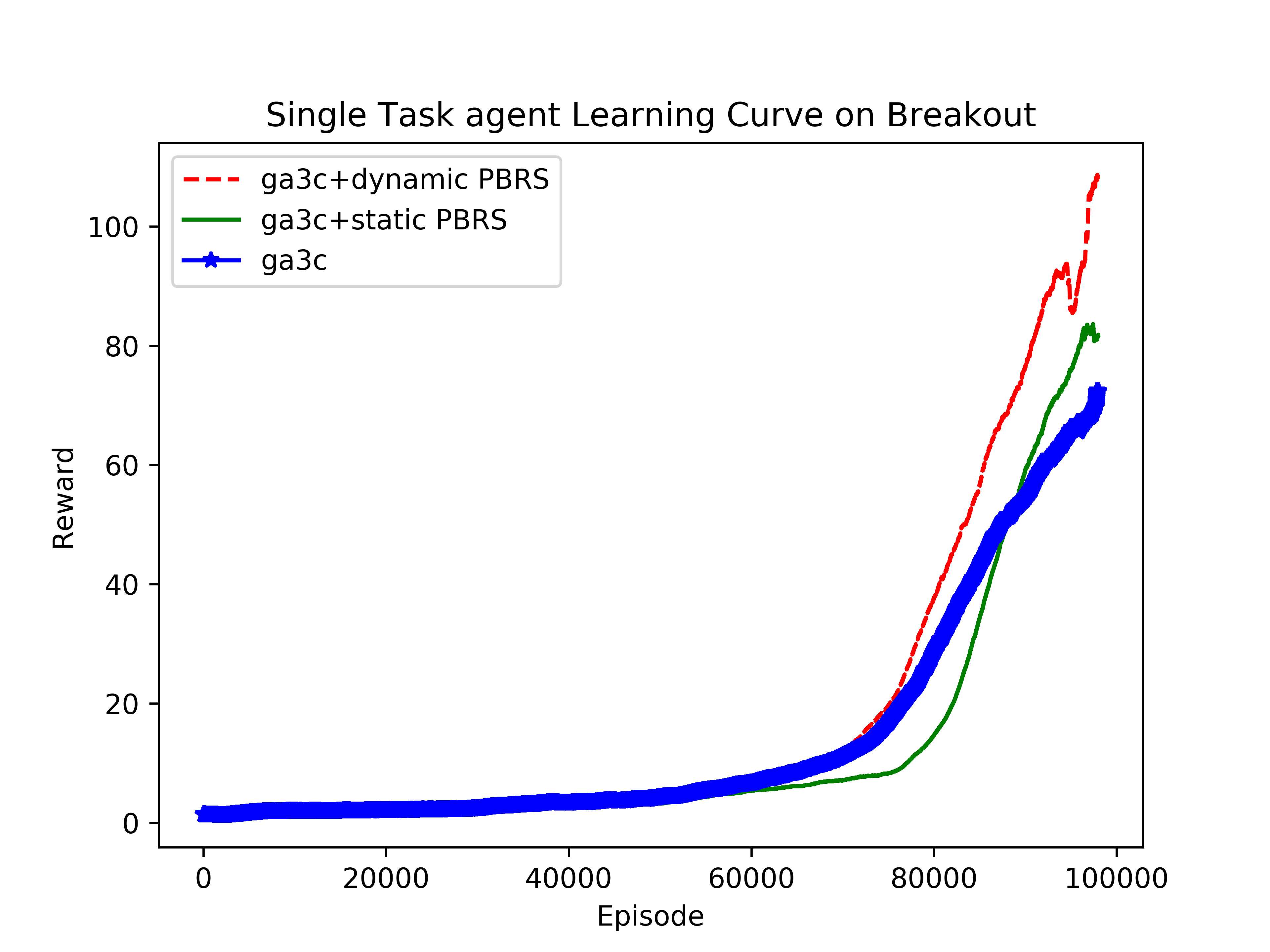}
        \caption{single task agent learning curve on Breakout}
\label{fig:breakout-b}
\end{figure}

\begin{figure*}[h]
\begin{centerline}
	\centering
	\begin{subfigure}{.5\textwidth}
		\centering
		\begin{center}
		\includegraphics[width=\textwidth, trim={.7cm .7cm .7cm .7cm},clip] {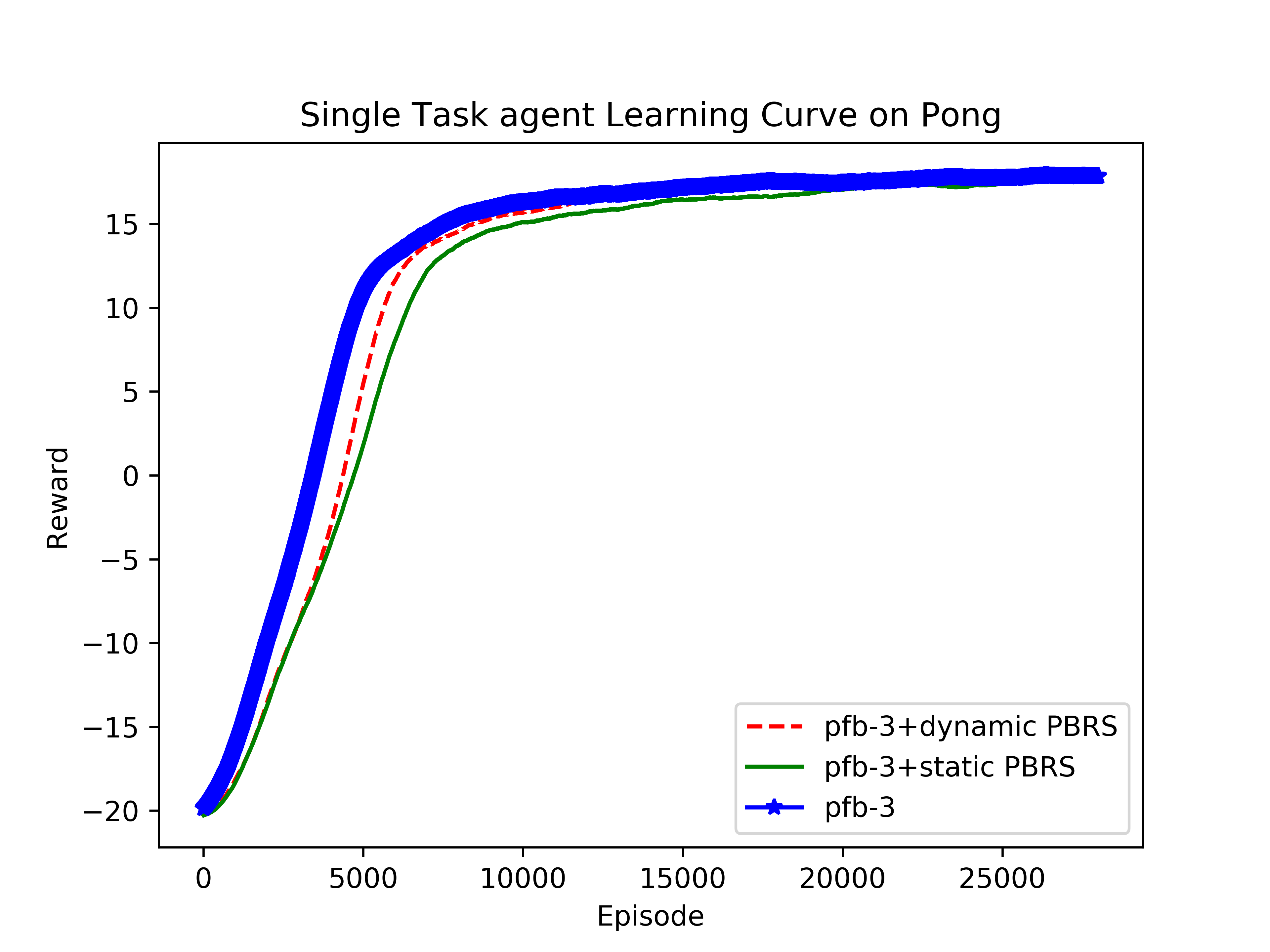}
		\caption{}
		\label{fig:pong-a}
		\end{center}
	\end{subfigure}
	\begin{subfigure}{.5\textwidth}
		\centering
		\begin{center}
		\includegraphics[width=\textwidth, trim={.7cm .7cm .7cm .7cm},clip]{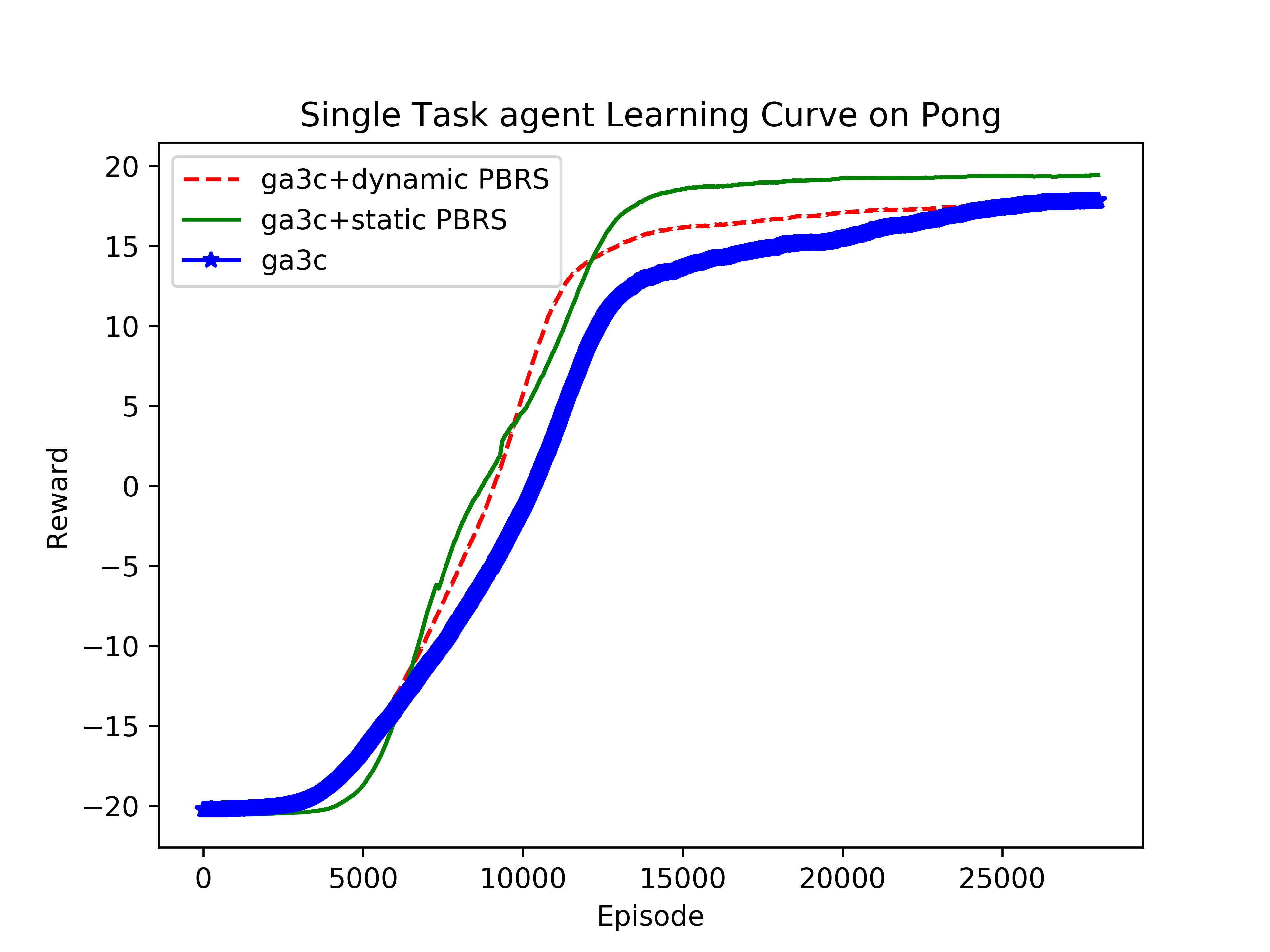}
		\caption{}
		\label{fig:pong-b}
		\end{center}
	\end{subfigure}
\caption{single task agent learning curve on Pong}
\label{fig:pong}
\end{centerline}
\end{figure*}

\textbf{How our proposed method works?}
By decomposition of equation \ref{eq:12} and considering equation \ref{eq:p-timeVar}, we can analyze how the proposed method works from a state to another state; the equation \ref{eq:16} is the result of this decomposition. During the learning process of an agent in an environment with sparse reward, three situations will appear by transmission from a state to another state: most of the time, the agent receives no reward and the episode reward will not change; sometimes the agent will receive a positive reward signal and it increases the episode reward; sometimes it receives a negative reward signal which causes a decrease in episode reward. 

First of all, let's assume that by transmission from a state to another state no change in episode reward has been made. So, as we can see in equation \ref{eq:17}, the potential function will be zero or negative. This will give the agent an information about the problem that means by remaining in the current state, it will not recieve any reward and it also may lose something. Hence, this negative feedback will force the agent to avoid remaining in a state and to make an improvement from a state to another state. The more the agent stays in a state, it will recieve more punishment. The more the agent in the vicinity of the goal and hesitate in a state, it will recieve more punishment. It will lead to more exploration near the target and avoid sticking to sub-optimal goals.

How about the case when there is an increase in the $episode\,reward$? If all of the future rewards are equally valuable, the potential function will return zero every time this situation appears to the agent. Whenever the episode reward has increased, as we can see in equation \ref{eq:18}, the value of the potential function will be positive, and this positive value reinforces the environment reward.

Another situation that the agent might face is a decrease in the episode reward. Decreasing in the episode reward might be greater than a threshold or less than a threshold. While this decline is greater than a threshold (equation \ref{eq:19}), the agent will receive a negative potential signal, which means the agent made a mistake or it spends its energy in a wrong way. Whenever the decline is less than the threshold (equation \ref{eq:20}), it will still receive a positive potential signal, which provides the flexibility of spending more energy on improving the agent performance in the way of achiving the goal.


\begin{figure*}[h]
	\centering
	\begin{subfigure}{.5\textwidth}
		\centering
		\begin{center}
		\includegraphics[width=\textwidth, trim={.7cm .7cm .7cm .7cm},clip]{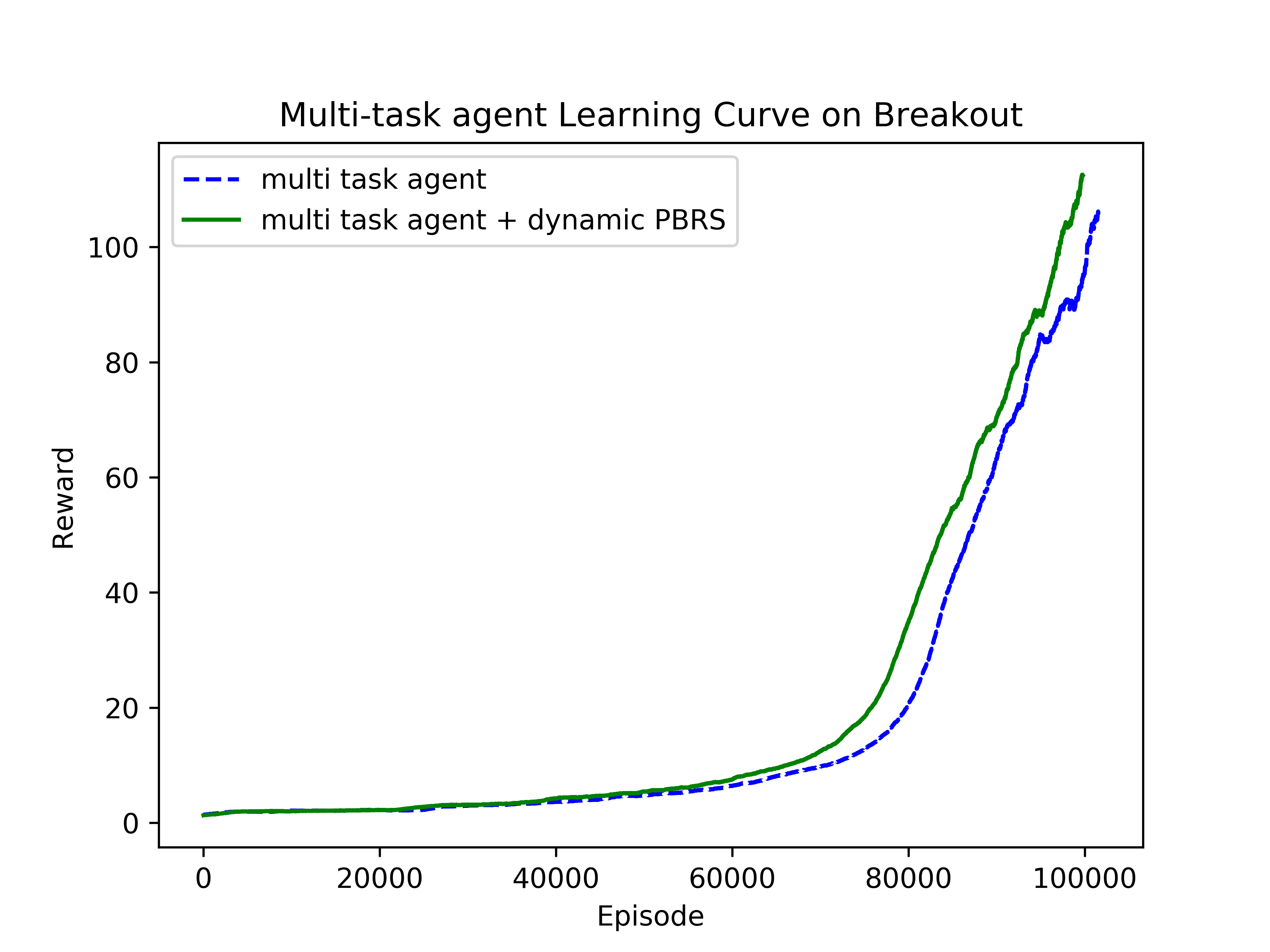}
		\caption{}
		\label{fig:ml-breakout}
		\end{center}
	\end{subfigure}%
	\begin{subfigure}{.5\textwidth}
		\centering
		\begin{center}
		\includegraphics[width=\textwidth, trim={.7cm .7cm .7cm .7cm},clip]{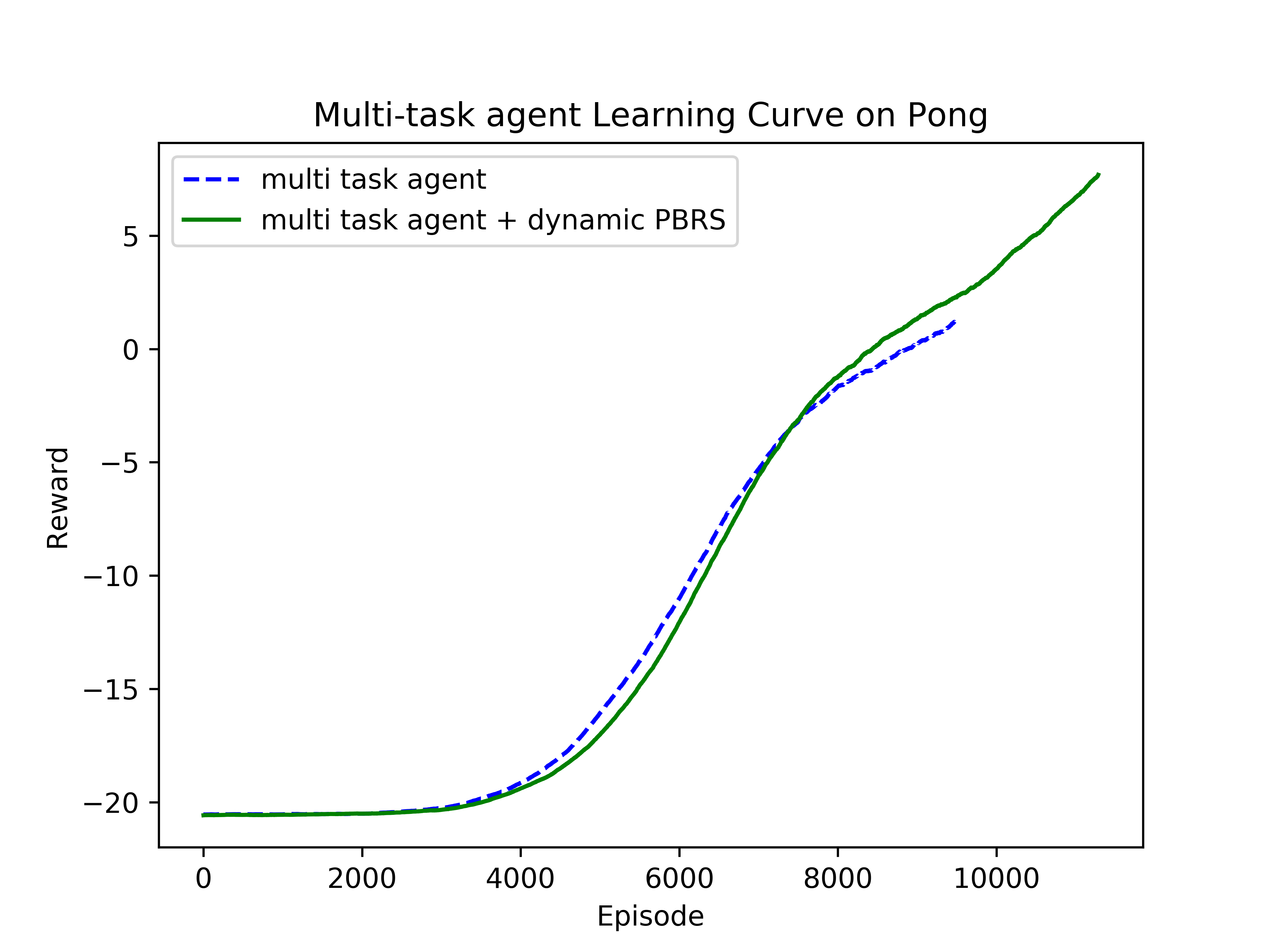}
		\caption{}
		\label{fig:ml-pong}
		\end{center}
	\end{subfigure}
\caption{multi-task agent learning curve on Pong and Breakout}
\label{fig:multi-learning}
\end{figure*}

\begin{figure*}[h]
\begin{centerline}
	\centering
	\begin{subfigure}{.5\textwidth}
		\centering
		\begin{center}
		\includegraphics[width=\textwidth, trim={.7cm .7cm .7cm .7cm},clip]{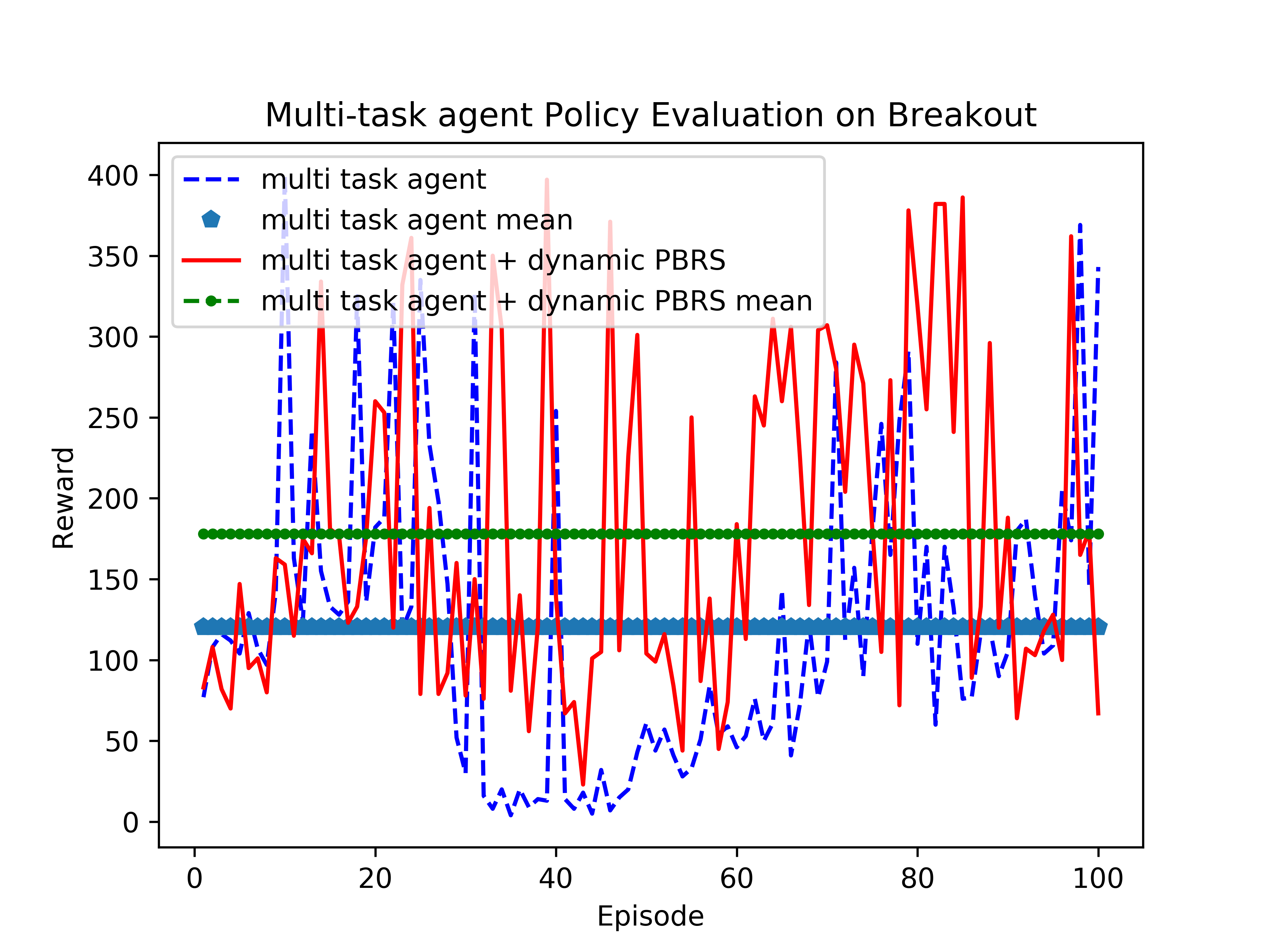}
		\caption{}
		\label{fig:mp-breakout}
		\end{center}
	\end{subfigure}
	\begin{subfigure}{.5\textwidth}
		\centering
		\begin{center}
		\includegraphics[width=\textwidth, trim={.7cm .7cm .7cm .7cm},clip]{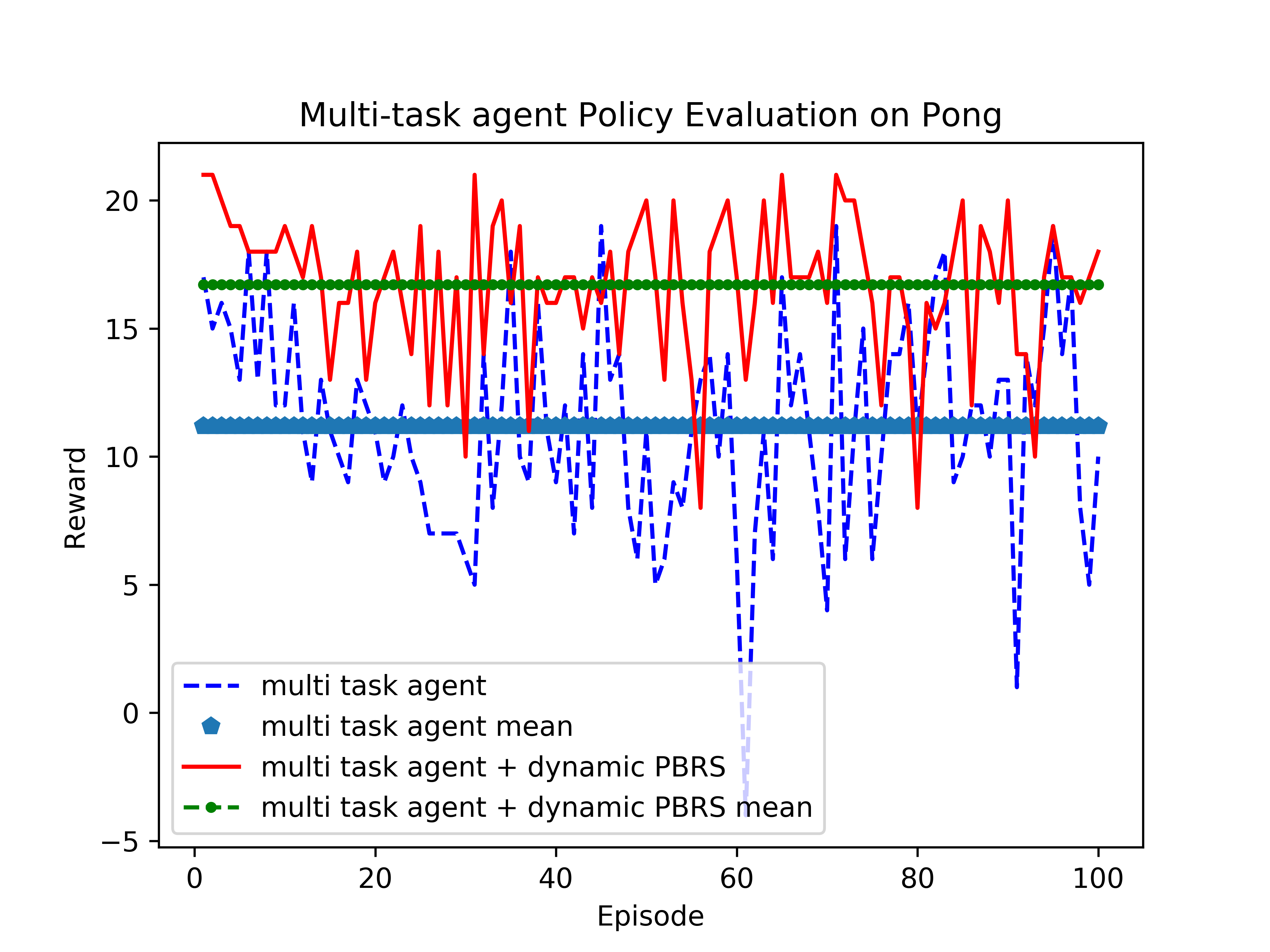}
		\caption{}
		\label{fig:mp-pong}
		\end{center}
	\end{subfigure}
\caption{multi-task agent policy evaluation on Pong and Breakout}
\label{fig:multi-policy}
\end{centerline}
\end{figure*}

\begin{align}
F(s,s^{'}) &= \gamma \phi (s^{'}) - \phi (s) \label{eq:12} \\
 &= \gamma - 1 + \gamma	\frac{R^{ep}(s^{'}) - R_{u}^{ep}(t)}{R_{u}^{ep}(t) - R_{l}^{ep}(t)} - 	\frac{R^{ep}(s) - R_{u}^{ep}(t)}{R_{u}^{ep}(t) - R_{l}^{ep}(t)} \label{eq:13} \\
  &= \gamma - 1 + \frac{\gamma R^{ep}(s^{'}) - R^{ep}(s) + R^{ep}_{u}(t) - \gamma R^{ep}_{u}(t)}{R_{u}^{ep}(t) - R_{l}^{ep}(t)}\label{eq:14} \\ 
  &= \gamma - 1 + \frac{\gamma R^{ep}(s^{'}) - R^{ep}(s) + R^{ep}_{u}(t) (1 - \gamma)}{R_{u}^{ep}(t) - R_{l}^{ep}}\label{eq:15} \\ 
    &= \frac{\gamma R^{ep}(s^{'}) - R^{ep}(s) + R^{ep}_{l}(t) (1 - \gamma)}{R_{u}^{ep}(t) - R_{l}^{ep}(t)}\label{eq:16}
  \end{align}

\begin{align}
    R^{ep}(s^{'}) = R^{ep}(s) &\Rightarrow F(s, s^{'})  =   \frac{(R^{ep}_{l}(t) - R^{ep} )(1 - \gamma) }{R_{u}^{ep}(t) - R_{l}^{ep}(t)} \le 0 \label{eq:17} \\
    \gamma R^{ep}(s^{'}) \geq R^{ep}(s) &\Rightarrow F(s, s^{'})  > 0 \label{eq:18}
\end{align}

\begin{align}
    \gamma R^{ep}(s^{'}) < R^{ep}(s) &\land \\
    \gamma R^{ep}(s^{'}) - R^{ep}(s) \, & >  -R^{ep}_{l}(t) (1 - \gamma) \Rightarrow  F(s, s^{'})  > 0 \label{eq:19}
\end{align}

\begin{equation}
\begin{split}
    \gamma R^{ep}(s^{'}) < R^{ep}(s) \, & \land \\
    \gamma R^{ep}(s^{'}) - R^{ep}(s) & \leq -R^{ep}_{l}(t) (1 - \gamma) \Rightarrow  F(s, s^{'})  \leq 0 \label{eq:20}
\end{split}
\end{equation}

\textbf{Extending to multi-task agents}
The presented potential function helps the learning methods to improve their performance by using knowledge extracted from previous episodes. Sometimes the agent needs to learn multiple tasks simultaneously. With this in mind, we extended the proposed method to be used in multi-task RL. According to the equation \ref{eq:p-multi}, which is an extension of \ref{eq:p-timeVar}, all parameters of the potential function need to be calculated separately for each of the tasks. By using equation \ref{eq:p-multi}, each of the tasks has a different potential function. So, the policy of each task remains optimal. In equation \ref{eq:p-multi}, the $i$ is an identifier for each of the tasks.
\begin{equation}\label{eq:p-multi}
\phi(s,a,t,i) = \Bigg\{ 
\begin{array}{ll}
	0 & R(s,a) = 0 \\
	1 + \frac{R^{ep}(i) - R_{u}^{ep}(t,i)}{R_{u}^{ep}(t,i) - R_{l}^{ep}(t,i)} & O.W
\end{array}
\end{equation}
\section{Results}
In order to demonstrate the usefulness of the proposed approach, we perform experiments in two domains: Breakout and Pong games from arcade learning environment (ALE) \cite{ale}. According to the \cite{ale}: 

``ALE provides an interface to hundreds of Atari 2600 game environments, each one different, interesting, and designed to be a challenge for human players. ALE presents significant research challenges for reinforcement learning, model learning, model-based planning, imitation learning, transfer learning, and intrinsic motivation."

We used two different baselines to compare our proposed method with them. \cite{ga3c} used as a first baseline and \cite{initial} as a second baseline. \cite{ga3c} is an implementation of \cite{a3c}, which uses a policy gradient method to approximate the optimal policy of a specific task. As a second baseline, we implement \cite{initial} that presented in the literature of transfer learning. We evaluate our work in two different phases: first of all, we evaluate our method during the learning process, and then we evaluate the performance of the final policy. We tested our method with two different assumptions: the first assumption is that we know the maximum and minimum values of the episode reward in each task, and the second assumption is that we have no information about the episode reward of the task and this values will be obtained during the learning process.

\textbf{Breakout:}
Breakout is an arcade game, which an agent handles a paddle to hit a ball trying to destroy more bricks while preventing the ball to cross the paddle. We trained an agent in this environment for 1000000 episodes. Figure \ref{fig:breakout-a} shows the learning curve of agents that trained on the Breakout game. As we see in figure \ref{fig:breakout-a} area under the curve of our method is greater than the baseline method, and the final performance is also better than the baseline.

\textbf{Pong:}
Pong is also an arcade game, which an agent has the responsibility of moving a paddle to hit a ball. The agent will receive -1 reward if it loses the ball and will receive +1 if the opponent loses the ball. We trained an agent in this environment for 30000 episodes. Figure \ref{fig:pong} shows the learning curve of agents that trained in the Pong game. As you see in figure \ref{fig:pong} area under the curve of our method is greater than one of the baseline methods. However, in general, the proposed method in this environment has not been able to work very well and has a little improvement.

\textbf{Multi-task Agent:}
We also implemented a multi-task agent, which learns how to play games of Pong and Breakout and then trains the agent for 115k episodes. We reinforced the reward signal using equation \ref{eq:p-multi}. Figure \ref{fig:multi-learning} and figure \ref{fig:multi-policy} demonstrate results of our experiments for a multi-task agent. Figure \ref{fig:multi-learning} demonstrates the learning curve of the multi-task agent for each of the games, while figure \ref{fig:multi-policy} demonstrates the performance of the learned-policy. Figure \ref{fig:multi-policy} is the result of running the agent with the learned-policy for 100 episode and then take the average rewards of episodes. As we can see, the multi-task agent, which using reinforced reward signal, does perform better on average than a multi-task agent, which does not use reinforced reward signal.

\section{Conclusion}
First of all, in this paper, we studied the literature on transfer learning and potential-based reward shaping.
According to this study, there was no method that tries to extract knowledge from the learning process. Hence, we introduced a novel way of extracting knowledge from the learning process, and we used reward shaping as a knowledge transferring method.

The proposed method guides the agent toward a goal by looking at the value of episode reward. If the agent goes toward the goal, the reward signal will be reinforced. This method can be used with any learning algorithm, and it supposed to be efficient wherever applied. We implemented our method using \cite{ga3c} and compare the results with two different baselines in two different environments, and then we extend our method to be used in multi-task agents. In most of the experiments, the results were promising. We also tested our method in a multi-task agent, which tries to learn games of Pong and Breakout simultaneously, and we saw that our method leads to improvement. 


{\small
\bibliographystyle{IEEEtran}
\bibliography{references}
}

\end{document}